\documentclass{article}
\usepackage{ijcai24}

\usepackage{times}
\usepackage{soul}
\usepackage{url}
\usepackage[utf8]{inputenc}
\usepackage[small]{caption}
\usepackage{graphicx}
\usepackage{amsmath}
\usepackage{amsthm}
\usepackage{booktabs}
\usepackage{algorithm}
\usepackage{algorithmic}
\usepackage[switch]{lineno}

\usepackage{enumitem}
\usepackage{zhlipsum}
\usepackage{colortbl}
\usepackage{amsfonts}
\usepackage{multirow} 

\usepackage[dvipsnames]{xcolor}

\definecolor{citecolor}{HTML}{0071BC}
\definecolor{linkcolor}{HTML}{ED1C24}

\usepackage[colorlinks,
            anchorcolor=red,
            citecolor=citecolor, 
            linkcolor=linkcolor,
            ]{hyperref}

\def\ie{\emph{i.e.,~}}
\def\eg{\emph{e.g.,~}}

\newcommand{\supp}{\emph{supplementary material}}

\newcommand{\NA}{---}

\ifdefined \GramaCheck
  \newcommand{\CheckRmv}[1]{}
  \newcommand{\figref}[1]{Figure 1}%
  \newcommand{\tabref}[1]{Table 1}%
  \newcommand{\secref}[1]{Section 1}
  \newcommand{\algref}[1]{Algorithm 1}
  \renewcommand{\eqref}[1]{Equation 1}
\else
  \newcommand{\CheckRmv}[1]{#1}
  \newcommand{\figref}[1]{Figure~\ref{#1}}
  \newcommand{\tabref}[1]{Table~\ref{#1}}
  \newcommand{\secref}[1]{Section~\ref{#1}}
  \renewcommand{\eqref}[1]{(\ref{#1})}
\fi

\newcommand{\sArt}{{state-of-the-art~}}

\newcommand{\zeroOneTwoThree}{{Zero-1-to-3}}

\title{Learning a Category-level Object Pose Estimator without Pose Annotations}

\author{
Fengrui Tian$^1$ \and
Yaoyao Liu$^2$ \and
Adam Kortylewski$^4$$^5$ \and
Yueqi Duan$^3$\and
Shaoyi Du$^1$\and
Alan Yuille$^2$\And
Angtian Wang$^2$\\
\affiliations
$^1$Xi'an Jiaotong Univeristy  
$^2$Johns Hopkins University  
$^3$Tsinghua University   \\
$^4$University of Freiburg
$^5$Max Planck Institute for Informatics \\
\emails
tianfr@stu.xjtu.edu.cn \quad
\{yliu538, ayuille1, angtianwang\}@jhu.edu \\
kortylew@cs.uni-freiburg.de \quad
duanyueqi@tsinghua.edu.cn \quad
dushaoyi@xjtu.edu.cn
}
\begin{document}

\maketitle

\begin{abstract}
% 3D object pose estimation is a challenging task. Previous works always require thousands of object images with annotated poses for learning the 3D pose correspondence, which is laborious and time-consuming for labeling. In this paper, we propose to learn a category-level object pose estimator without pose annotations. 

% Specifically, we learn object-pose discriminability from controlled novel view generation of a set of unannotated images using diffusion models (\ie~\zeroOneTwoThree).

% Although the diffusion models create images with rich but crude object pose information, the pose control is rather noisy while the generated images may have significant artifacts of image qualities (\eg incorrect textures and limited resolution).

% To alleviate the quality issue and pose noise, we propose to build 3D pose correspondence based on image features with a coarse-level object geometry. 

% Specifically, we exploit an image encoder, which is learned from a specially designed contrastive pose learning, to filter the unreasonable details and extract image feature maps. 
% Additionally, we propose a novel learning strategy that allows the model to learn object pose from those generated image sets without knowing the alignment of their canonical poses.

% Experimental results show that our method has capability of category-level object pose estimation from a single shot setting (as pose definition), while significantly outperforming other \sArt~methods on the few-shot category-level object pose estimation benchmarks.

3D object pose estimation is a challenging task. Previous works always require thousands of object images with annotated poses for learning the 3D pose correspondence, which is laborious and time-consuming for labeling. In this paper, we propose to learn a category-level 3D object pose estimator without pose annotations. Instead of using manually annotated images, we leverage diffusion models (\eg~\zeroOneTwoThree) to generate a set of images under controlled pose differences and propose to learn our object pose estimator with those images. Directly using the original diffusion model leads to images with noisy poses and artifacts. To tackle this issue, firstly, we exploit an image encoder, which is learned from a specially designed contrastive pose learning, to filter the unreasonable details and extract image feature maps. Additionally, we propose a novel learning strategy that allows the model to learn object poses from those generated image sets without knowing the alignment of their canonical poses. Experimental results show that our method has the capability of category-level object pose estimation from a single shot setting (as pose definition), while significantly outperforming other \sArt~methods on the few-shot category-level object pose estimation benchmarks.

% \angtian{In contrast to previous object pose estimation approaches, which require hundreds or thousands of pose annotated images for training. Furthermore, annotating 3D object poses is far more time-consuming than labeling for class or bounding boxes. This extensive requirement for human labor impedes the pathway to real-world applications, which demand the annotation of millions of objects.}
% Following the philosophy of \textit{Analysis-by-Synthesis}, previous works mostly synthesize a category-level object representation and analyze the pose of a novel instance by comparing the object image with the synthesized representation. As a result, they need to annotate the object poses on the hundreds or thousands of images to train the category-level representation, which needs lengthy annotating time when applying to millions of objects in real-world applications. To address this, we propose to learn a category-level object estimator by jointly optimizing hundreds of instance-level object representations. More specifically, we exploit 3D object generation model (\ie~\zeroOneTwoThree~\cite{23iccv/liu_zero123}) to generate multiple views of each object, and conduct a unified image encoder to extract image feature maps of the object views. We separately introduce a neural mesh representation for each instance and optimize all the instance-level representations simultaneously.
\end{abstract}

\section{Introduction}
\label{sec:intro}
\begin{figure}[t]
  \centering
%   \fbox{\rule{0pt}{2in} \rule{0.9\linewidth}{0pt}}
   \includegraphics[width=0.97\linewidth]{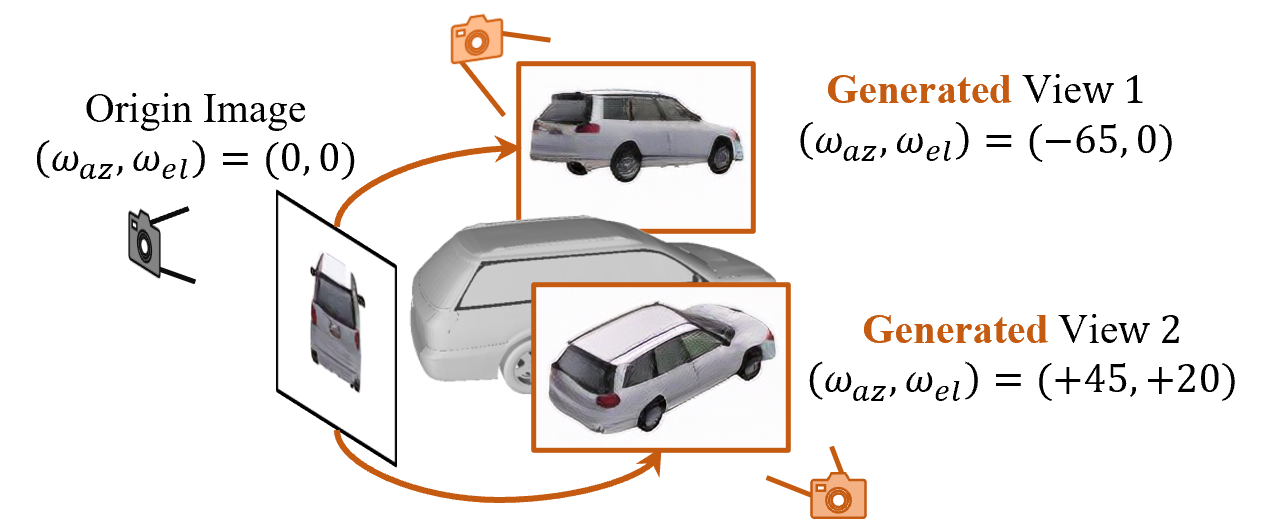}
   \caption{We propose to learn a category-level object pose estimator from the multiple views of the objects in the category. As shown in the figure, by leveraging the generative diffusion model, we generate novel views of an object with controlled poses $(\omega_{az}, \omega_{el})$ and learn the pose estimator from the multiple views of the objects in the category and their controlled poses.}
   \label{fig:teaser}
\end{figure}
The 3DoF rotation of objects is an inherent characteristic of real-world objects. 
% The 3D object pose, in turn, has a profound impact on the visual appearance of these objects, ensuring image consistency.
Estimation of 3D poses of objects offers advantages for downstream applications, such as autonomous driving and robotics.  
% On the other hand, the inverse task \ie estimating 3D object poses also offers advantages for downstream applications, such as autonomous driving and robotics.  
Recent researchers \cite{21nips/wang_nvsm,21iclr/wang_nemo,22cvpr/yin_fishermatch} tried to build category-level estimators to estimate object poses from a particular category. One popular solution is following the \textit{Analysis-by-Synthesis} principle. Specifically, given a set of object images with annotated poses, these methods first build a 3D neural mesh as the category-level object representation and analyze the pose of a novel object by comparing the 2D image of the object with the 3D mesh. 
In this way, the category-level generalization ability of the model could be learned from thousands of annotated images. 
Based on the category-level representation, previous works devote efforts to shape prior \cite{wang2024neural}, occlusion problem \cite{ma2022robust} and keypoint detection \cite{15cvpr/tulsiani_view} to learn a more robust category-level pose estimator.

However, these category-level pose estimation methods need hundreds of annotated images for a novel object class to learn a unified representation. Although some researchers \cite{21nips/wang_nvsm,22cvpr/yin_fishermatch} proposed to learn object pose estimators with fewer annotated images, those approaches still require a certain amount of human labor on annotation,  which impedes the pathway to millions of object categories in the real world. A natural question is raised: can we learn a category-level object pose estimator without pose annotations?

% Recently a series of diffusion-based 3D generative models \cite{23iccv/liu_zero123,23arxiv/shi_zero123plus} show an ability to recover the 3D geometry from a single image with viewing controls. Specifically, given an object image and the target viewing pose, these models output an image containing the object at the target view. 
% Although the generated object images at different views may suffer from the limited image qualities and inaccurate controls of viewing angles, such models demonstrate great potential in many downstream tasks \cite{23cvpr/ruiz_dreambooth,23cvpr/kawar_imagic,23cvpr/zhang_invstyle}.

In this paper, we propose an affirmative answer to this question. 
Instead of using annotated images to build category-level representations, we learn our object pose estimator from image sets that each image set is generated from a single unannotated image using diffusion models. 
Built on top of the recent series of diffusion-based 3D generative models \cite{23iccv/liu_zero123,23arxiv/shi_zero123plus}, which show the ability to recover the 3D geometry from a single image with viewing controls, we generate the set of images under a controlled pose different from the set of original unannotated images for training.
However, we find that the generated images have some artifacts for the image quality, \eg~incorrect texture, and the pose control is rather crude, \ie~noise for the pseudo pose label of the generated images.
To resolve this, we introduce an image encoder as an information filter for removing unreasonable object details, and a merging strategy that distillates the learned pose discriminability from all learned object representations.

As shown in \figref{fig:teaser}, given a set of object images with target view poses, we first exploit a generative diffusion model (\ie~\zeroOneTwoThree~\cite{23iccv/liu_zero123}) to generate a set of images in a certain pose difference of the object in the original image. We introduce an image encoder for filtering inconsistent details and extracting the feature maps of the generated images. Then we build a separate neural mesh for each instance and jointly optimize all the neural meshes with the image feature maps. 
% Since we do not know the object poses and cannot align the mesh shape to the instances on the images, we select the unit sphere as the mesh shape and optimize all the instance representations simultaneously. 
% As shown in \fengrui{Figure: teaser}, our model is able to predict the object pose by only using several unannotated images from the Internet. 
% Moreover, in the traditional few-shot settings, experimental results on the PASCAL3D+ dataset demonstrate that our model outperforms other \sArt methods by a large margin.
The experiments conducted on the PASCAL3D+ \cite{14wacv/xiang_pascal3d} and KITTI \cite{13ijrr/geiger_kitti} datasets demonstrate that our approach is capable of performing object pose estimation using a single shot annotation as the definition of pose.
Moreover, we compare our approach with the \sArt methods for the few-shot category-level 3D object pose estimation on the PASCAL3D+ dataset. The experimental results demonstrate that our model outperforms previous \sArt methods by a large margin.

In summary, our contribution includes:
\begin{itemize}
    \item We propose a category-level object pose estimation pipeline that learns 3D object poses without requiring the need for pose annotations.
    \item We introduce a diffusion-based image generation pipeline, which creates pose-controlled images from a single unannotated image.
    % We propose to learn a category-level object pose estimator from multiple views of hundreds of objects in the category, without requiring the need for pose annotations.
    \item We propose a learning pipeline, which learns object-pose discriminability from multiple generated image sets jointly and combines all the learned instance-specific estimators.
    % We build the 3D pose correspondence from the extracted feature maps to alleviate the negative effect of Janus problems on the generated view images of objects.
    \item Experimental results show that our model could predict object poses in a new category by training with unannotated images and achieves \sArt~performance on the few-shot pose estimation setting.
\end{itemize}

\section{Related Work}
\label{sec:related_work}
\begin{figure}[t]
  \centering
%   \fbox{\rule{0pt}{2in} \rule{0.9\linewidth}{0pt}}
   \includegraphics[width=0.97\linewidth]{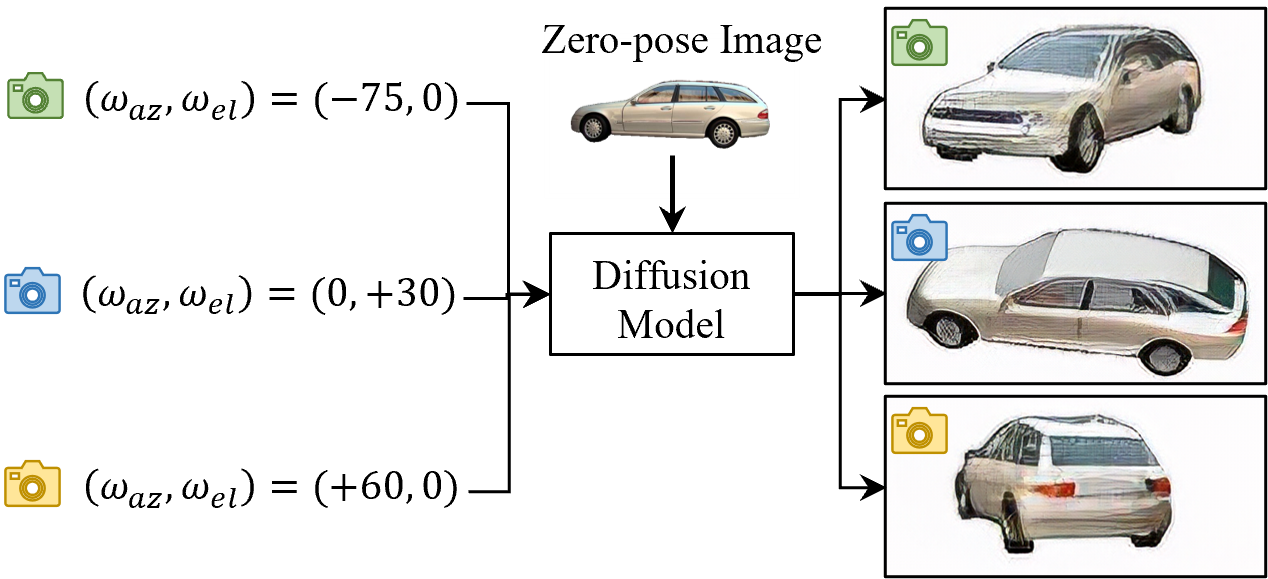}

   \caption{Posed image generation. Given an image containing an object and the pose of a target view $(\omega_{az}, \omega_{el})$, we control the diffusion model to generate the target view of the object by leveraging the view pose.}
   \label{fig:view_gen}
\end{figure}
\paragraph{Learning a pose estimator with full supervision.}
3D object pose estimation is a challenging task. Early works either considered the pose estimation challenge as a regression problem \cite{15cvpr/tulsiani_view,17cvpr/mousavian_3d}, or studied the problem as a two-step approach \ie keypoint detection and Perspective-n-Point solving process \cite{09ijcv/lepetit_epnp,17icra/pavlakos_6dof,18eccv/shou_starmap}.
Recently a series of works tried to solve the problem by following the Analysis-by-Synthesis philosophy \cite{19cvpr/wang_norm,20iros/lin_inerf}. Specifically, they used a differentiable renderer to render a synthesized image and conduct pose estimation by minimizing the reconstruction loss between the synthesized image and the target image. Such "render-and-compare" pipeline has been extended by NeMo \cite{21iclr/wang_nemo}, in which the comparison is conducted between the extracted feature banks of the synthesized image and target image. However, these approaches need a huge amount of pose annotations, which are time-consuming and laborious for human labeling. In this study, we tried to learn the pose estimator without pose annotations.
\begin{figure*}[!t]
  \centering
%   \fbox{\rule{0pt}{2in} \rule{0.9\linewidth}{0pt}}
   \includegraphics[width=0.97\linewidth]{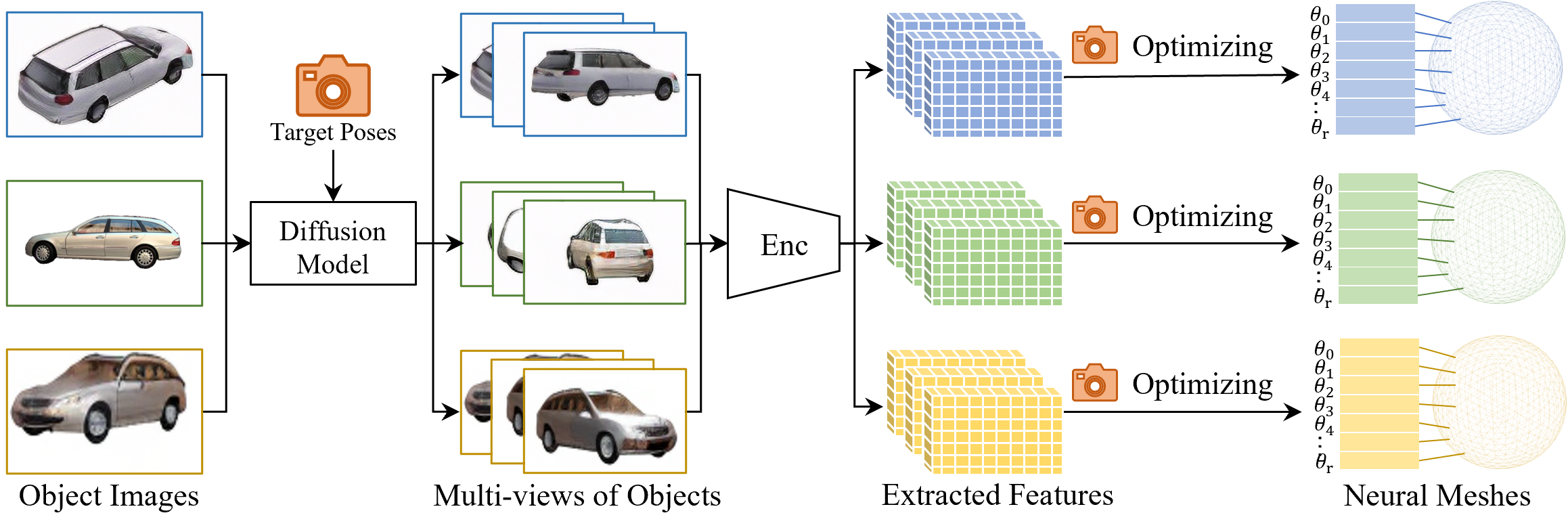}

   \caption{The training pipeline of our model. Given a set of object images, we define the object poses on these images as zero-poses (\ie $(\omega_{az}, \omega_{el})=(0,0)$). Then given the target view poses, we exploit the generative diffusion model to generate the target views of the objects. We introduce an image encoder to extract the image feature maps of these view images. We exploit the image feature maps with the corresponding target poses to optimize the neural mesh for each object.}
   \label{fig:overview}
\end{figure*}
\begin{figure}[t]
  \centering
%   \fbox{\rule{0pt}{2in} \rule{0.9\linewidth}{0pt}}
   \includegraphics[width=0.97\linewidth]{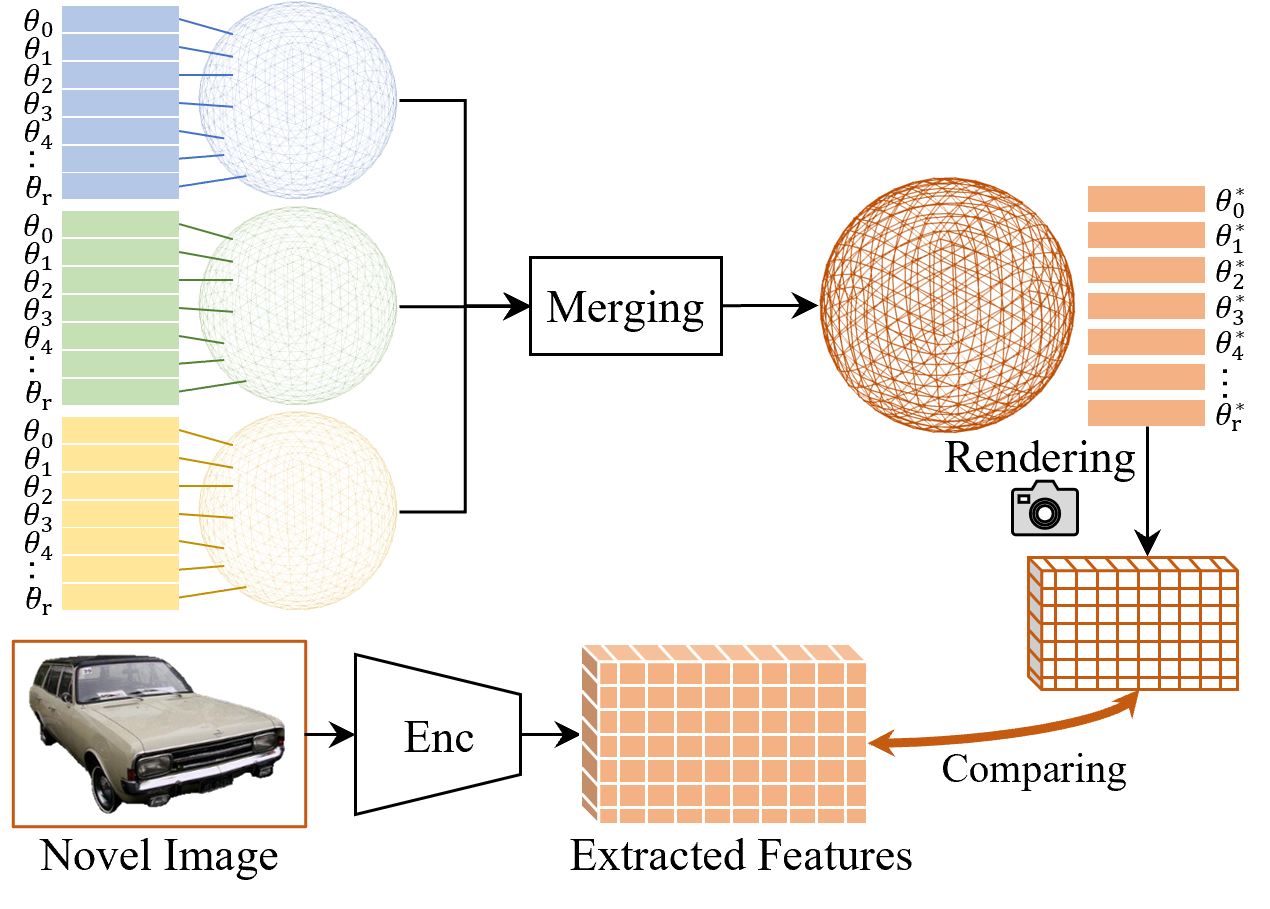}

   \caption{Neural mesh merging strategy and evaluation pipeline. After training, we estimate the relative pose between two meshes and merge two neural meshes with the estimated pose. In the evaluation, We consider the pose of a novel image as a learnable matrix. We render the feature map from the merged neural mesh with the learnable pose and optimize the pose by comparing the rendered feature map with the extracted feature map of the novel image.}
   \label{fig:overview_eval}
\end{figure}
\paragraph{Learning a pose estimator in the few-shot setting.} To reduce the consumption of annotated data, recent studies focused on the pose estimation problem in a few-shot setting \cite{21nips/wang_nvsm,22cvpr/yin_fishermatch,22iclr/wang_voge,23arxiv/jesslen_robust,24wacv/yang_robust}. For example, NVSM \cite{21nips/wang_nvsm} proposed to learn an object pose estimator with few annotated images and a collection of unlabeled data. FisherMatch \cite{22cvpr/yin_fishermatch} proposed a teacher-student framework that facilitates the information flow from annotated data to unlabeled data. Although these methods successfully reduce the need for annotated data by training with hundreds of unannotated images, they still require a certain amount of annotated data for each object category, which significantly limits its further applications in the real world when encountering billions of objects in millions of categories. In this study, we try to train the estimator without the need for any pose annotations.
% \subsection{Category-level Object Pose Estimation}

\paragraph{3D object generation.}

Recently, the diffusion models \cite{20nips/ho_ddpm,20iclr/song_ddim} bring a storm on the generation tasks \cite{22iclr/poole_dreamfusion,23iccv/liu_zero123,23arxiv/shi_zero123plus,23cvpr/lin_magic3d,23nips/liu_one}. The images generated from these models contribute to many downstream applications such as text-to-image generation \cite{23cvpr/ruiz_dreambooth,24tvcg/gao_guess}, image editing \cite{23cvpr/kawar_imagic} and image style transformation \cite{23cvpr/zhang_invstyle}. The high quality of 2D-generated images facilitates many researchers to study the 3D object generation. \zeroOneTwoThree~\cite{23iccv/liu_zero123} is a representative one that exploits an object image to generate the 3D reconstruction of the object. It uses rotation angles to control the generated view of the object. Although its following works such as \zeroOneTwoThree++ \cite{23arxiv/shi_zero123plus} improve the image generation quality, the generated views of objects still encounter the image quality issues and inaccurate poses of the generated objects, which hinder the further applications to the real world. In this paper, we leverage the power of the generative diffusion model and propose to learn from the encoded features of the generative view images to reduce the negative impacts.

\section{Method}

In this section, we introduce our method for learning a category-level object pose estimator without pose annotations. Specifically, given a set of unannotated images $\mathcal{I}$, for each image $I_y \in \mathcal{I}$, we first define its canonical pose space center at $P_y$, where $P_y$ is unknown during training. In \secref{sec:image_generation} we introduce how to generate a novel view of an object image under a certain rotation $\Delta P_m$. Then we introduce the learning strategy by leveraging the generated views of one object image in \secref{sec:specific}. and describe the overall training strategy in \secref{sec:joint_learn}. We finally present the evaluation pipeline in \secref{sec:eval}. 

\subsection{Posed Image Generation from Diffusion}
\label{sec:image_generation}

% \todo{discuss the definition of object pose}
% \todo{discuss the definition of related pose and canonical pose of object}
% \todo{discuss given a pose space A, how to convert it into pose space B.}

As \figref{fig:overview} shows, we learn our category-level object pose estimator from the image sets generated by the diffusion model. Each image set $\mathcal{I}_y = \{ I_m \in \mathbb{R} ^ {H \times W \times 3} \}$ is generated from a real image $\hat{I}_y$ in the training set.
Specifically, we first exploit a segmentation method (\ie~SAM \cite{23arxiv/kirillov_sam}) to delineate the object from the image. For each delineated object image, we define its canonical pose space center $P_y \in \mathbb{R}^{3\times3}$ as the object pose on the image. Since we do not know the pose of the object, we initialize $P_y$ as an identity matrix. Then as shown in \figref{fig:view_gen}, given the rotation angle $(\omega_{az}, \omega_{el})$ of a target view, we use a generative diffusion model (\ie~\zeroOneTwoThree) to generate the object image viewing from the target direction,
\begin{equation}
    I_m = \Psi_{gen}(\hat{I}_y, (\omega_{az}, \omega_{el})),
\end{equation}
where $I_m$ is the generated object image. 
% In the object-centric coordinate system, the rotation of viewing could be transformed into the rotation of the object. 
By assuming the object facing toward the viewing direction of the camera coordinate of the image, each pose of the generated image can be described as a composition of canonical pose $P_y$ with an additional rotation $\Delta P_{m}$. The $\Delta P_{m}$ is computed by
% Concretely, let $T$ denote the transformation function, we can obtain the target view image by rotating the object with rotation matrix $\Delta P_{m}$,
\begin{equation}
    \Delta P_{m} = T(\omega_{az}, \omega_{el}),
\end{equation}
where the $T$ denotes a transformation function (known as the LookAt transformation).
Then the rotated object pose of $I_m$ could be defined by the following equation,
\begin{equation}
    P_m = P_{y}\Delta P_{m}.
\end{equation}
Summarily, we obtain an image set $\mathcal{I}_y$ of the object on the image ${I}_y$ containing the generated object images from different angles and the pose set $\mathcal{P}_y$ containing the corresponding image poses.
Due to the limited resolution of the generated image, we exploit the super-resolution model \cite{18eccvw/wang_esrgan} to upsample the image from 256$\times$256 to 512$\times$512. 
% In the experiment, we also observe a significant image quality improvement with more details when utilizing the super-resolution model.

\subsection{Canonical Specific Object Pose Learning}
\label{sec:specific}
We formulate our object pose estimator following Neural Meshes Models \cite{21iclr/wang_nemo}. Specifically, for the image set $\mathcal{I}_y$, we learn a neural mesh $\mathfrak{N}_y=\{\mathcal{V}_y, \mathcal{A}_y, \Theta_y\}$, which consists a set of mesh vertices $\mathcal{V}_y = \{V_k \in \mathbb{R}^3\}_{k=1}^{K_y}$, triangular faces $\mathcal{A}_y = \{A_k \in \mathbb{N}^3\}_{k=1}^{K^{'}_y}$ and feature vectors on each vertex $\Theta_y = \{\theta_k \in \mathbb{R} ^ d\}_{k=1}^{K_y}$, where $K_y$ and $K^{'}_y$ are the numbers of vertices and faces of each mesh. Different from the uniformly sampled or cuboid object meshes used in NeMo \cite{21iclr/wang_nemo}, we exploit the geodesic polyhedrons as the object geometry representations, which allow to learn the neural meshes without the assumption of the canonical pose of each instance, since the geometry of geodesic polyhedrons is independent of the pose. 

In the training, we first use a ResNet feature extractor \cite{16cvpr/he_resnet} to extract the feature map $\Phi_w(I_m) = F \in \mathbb{R}^{c \times h\times w}$ from the input image $I_m \in \mathcal{I}_y$, where $\Phi_w$ is the feature extractor with network parameters $w$.
Then, to learn the vertex features $\theta_k$, as well as the feature extractor $\Phi_w$, we calculate the world-to-screen transformation $\Omega_m$ given a camera pose $P_m \in \mathbb{R}^3$. 
To find the vertex $k$ corresponding feature $f_k = F(p_k)$ at pixel $p_k$ on the feature map, we compute the projected location of each vertex on the feature map $p_k=\Omega_m(V_k)$.
Besides, the visibility $o_k$ is determined for each vertex in the image, \ie $o_k=1$ if vertex $k$ is visible, and vice versa. Specifically, we render the depth map $\mathbf{D}=\textit{Render}(\mathfrak{N}_y, \Omega_m)$ and compute the vertex-to-camera distance $\mathbf{d}_k = \Vert Q - V_k \Vert_2$. Then the vertex visibility is computed as 
\begin{equation}
    o_k=
    \begin{cases}
      0, \Vert \mathbf{D}_{p_k} - \mathbf{d}_k \Vert_2  > \tau_r \\
      1, \Vert \mathbf{D}_{p_k} - \mathbf{d}_k \Vert_2  \leq \tau_r
    \end{cases},
\end{equation}
where $\tau_r$ is a preset threshold.
We learn the vertex features $\theta_k$ using the momentum update strategy~\cite{bai2020coke},
\begin{equation}
    \theta_k \xleftarrow{} o_k (1 - \beta) \cdot f_k + (1 - o_k + \beta \cdot o_k) \theta_k,
\end{equation}
where $\beta$ is the momentum for the update process.

% A contrastive loss is used to learn the weight $w$ in the feature extractor. 
Our learning optimal of the feature extractor is to enlarge the feature distance $\Vert \theta_j - f_i \Vert_2$ if $\Vert V_i - V_j \Vert_2$ is above a desired threshold. Such properties of features allow us to use differentiable rendering to find the optimal alignment of the vertices on the 3D model and corresponding locations on the 2D image. To achieve this, we use the contrastive loss~\cite{bai2020coke} to learn the weights $w$:
\begin{equation}
\label{eq:feat_loss}
     \mathcal{L}_{m}^{fg} = -\sum_k o_k \cdot \log(\frac{e^{\kappa f_k \cdot \theta_k}}{\sum_{\substack{\theta_l \in \mathcal{Y}_y, v_l\notin \mathcal{N}_k}} e^{\kappa f_k \cdot \theta_l} } ),
\end{equation}
where $\kappa$ is a preset softmax temperature, and $\mathcal{N}_k$ indicates a spacial neighborhood of $V_k$, which controls specially accuracy of the learned features. We also include the background loss $L_y^{bg}$ following~\cite{bai2020coke}. During training, we jointly optimize all the neural meshes $\{ \mathfrak{N}_1, \mathfrak{N}_2, ..., \mathfrak{N}_Y \}$ where $Y$ is the number of image instances.

\begin{figure*}[!t]
  \centering
%   \fbox{\rule{0pt}{2in} \rule{0.9\linewidth}{0pt}}
   \includegraphics[width=0.97\linewidth]{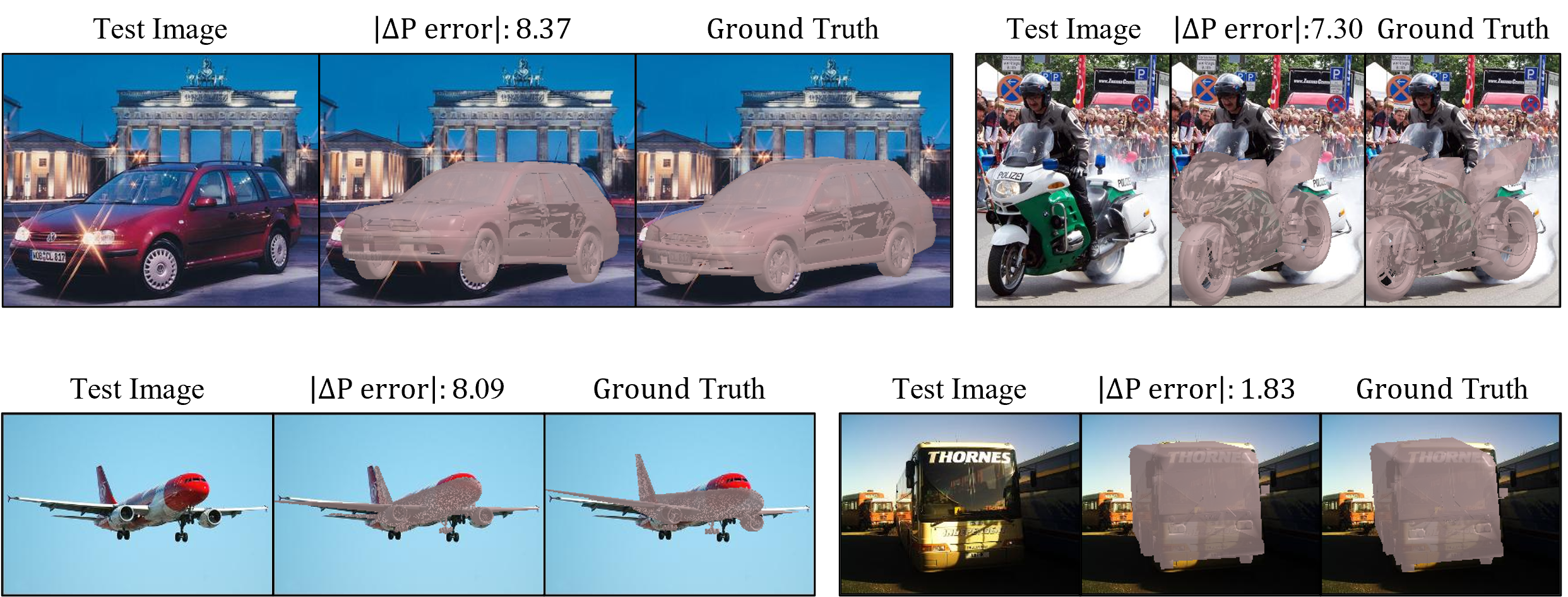}

   \caption{Qualitative results of training the pose estimator without pose annotations. We present four cases on the car, motorbike, aeroplane and bus categories, respectively. We use the CAD model of each category for better visualization and present the pose estimation errors on the top of the prediction images. Without requiring any pose annotations, our model still successfully predicts object poses in low pose estimation errors.}
   \label{fig:qualitative_wo_pose}
\end{figure*}

\subsection{Training Strategy}
\label{sec:joint_learn}
\newcommand{\supervise}[1]{{\color{gray} #1}}
\begin{table*}[]
\setlength{\tabcolsep}{4.5pt}
\centering
\begin{tabular}{l|cc|cc|cc|cc}
\toprule
\multirow{2}{*}{Method} & \multicolumn{2}{c|}{1} & \multicolumn{2}{c|}{7} & \multicolumn{2}{c|}{20} & \multicolumn{2}{c}{50} \\
 & Med. $\downarrow$ & Acc$_{30} \uparrow$ & Med. $\downarrow$ & Acc$_{30} \uparrow$ & Med. $\downarrow$ & Acc$_{30} \uparrow$ & Med. $\downarrow$ & Acc$_{30} \uparrow$ \\ \midrule 
Res50-Generic & \NA & \NA & 39.1 & 36.1 & 26.3 & 45.2 & 20.2 & 54.6 \\
Res50-Specific & \NA & \NA & 46.5 & 29.6 & 29.4 & 42.8 & 23.0 & 50.4 \\
StarMap \cite{18eccv/shou_starmap} & \NA & \NA & 49.6 & 30.7 & 46.4 & 35.6 & 27.9 & 53.8 \\
NeMo \cite{21iclr/wang_nemo} & \NA & \NA & 60.0 & 38.4 & 33.3 & 51.7 & 22.1 & 69.3 \\
NVSM \cite{21nips/wang_nvsm} & \NA & \NA & 37.5 & 53.8 & 28.7 & 61.7 & 24.2 & 65.6 \\
FisherMatch \cite{22cvpr/yin_fishermatch} & \NA & \NA & 28.3 & 56.8 & 23.8 & 63.6 & 16.1 & 75.7 \\
Ours  & \textbf{29.3} & \textbf{58.8} & \textbf{26.3} & \textbf{60.7} & \textbf{22.8} &\textbf{ 69.1} & \textbf{15.4} & \textbf{77.4 }\\
\midrule
\supervise{Full Supervision \cite{21iclr/wang_nemo}} & \supervise{8.1} & \supervise{89.6} & \supervise{8.1} & \supervise{89.6} & \supervise{8.1} & \supervise{89.6} & \supervise{8.1} & \supervise{89.6} \\
\bottomrule
\end{tabular} 
\caption{Few-shot pose estimation results on the PASCAL3D+ dataset. We indicate the number of annotations during training for each category and evaluate all the approaches using Accuracy (in percent, higher better) and Median Error (in degree, lower better). Notably, "1 instance" denotes that we train our model without annotations and only exploit one pose annotation on an image as the pose definition on the evaluation set. We also present the results from the fully supervised model at the bottom line. It can be seen that our model outperforms other \sArt~methods in the few-shot settings. Moreover, even without annotation images in the training steps, the Acc$_{30}$ performance of our model is still better than other methods that need seven annotated images for training.
%to provide context. Note the accuracy under full supervision is not comparable, since few-shot pose estimation is significantly harder.
}
\label{tab:pascal3d}
\end{table*}

\paragraph{Optimization loss.}
As we describe how to optimize a neural mesh with an input image in \secref{sec:specific}, now we introduce how to train the image encoder by jointly optimizing multiple neural meshes. For the generated images $\mathcal{I}_m$, we compute the feature loss $\mathcal{L}_m$ by exploiting \eqref{eq:feat_loss}. The final loss could be formulated as 
\begin{equation}
    \mathcal{L}_{\text{train}} = \sum_{\mathcal{I}_y \in \Im} \mathcal{L}_y^{\text{train}} = \sum_{\mathcal{I}_y \in \Im} \sum_{I_m \in \mathcal{I}_y}(\mathcal{L}^{fg}_m + \mathcal{L}^{bg}_m).
\end{equation}

\paragraph{Instance Neural Mesh merging via canonicalization.} 
We suppose that during training, each neural mesh learns a specific representation related to the corresponding object. Hence we propose a mechanism to merge the well-trained meshes into one category-level object representation.
% In the training stage, multiple neural meshes $\{ \mathfrak{N}_1, \mathfrak{N}_2, ..., \mathfrak{N}_Y \}$ are learned. 
% we propose a mechanism to merge the well-trained meshes into one category-level object representation once the training is done. 
As shown \figref{fig:overview_eval}, although we do not know the relative ground truth poses between different meshes, we could estimate the relative poses by exploiting the learned feature vectors. Specifically, given two neural meshes $\mathfrak{N}_i=\{\mathcal{V}_i, \mathcal{A}_i, \Theta_i\}, \mathfrak{N}_j=\{\mathcal{V}_j, \mathcal{A}_j, \Theta_j\}$, we define that the mesh $\mathfrak{N}_j$ is the anchor mesh that we always try to merge other meshes into the mesh $\mathfrak{N}_j$. We further define a rotation matrix $R$ that rotates the neural mesh $\mathfrak{N}_i$ at the target angle.
Then we could calculate the vertex distance between the vertex ${V}_m$ in $\mathcal{V}_{j}$ from mesh $\mathfrak{N}_j$ and the rotated vertex from the mesh $\mathfrak{N}_i$. 
\begin{equation}
    D_{mn}^{vertex}(R) =|{V}_{m} - R V_n|,
\end{equation}
where $V_n \in \mathcal{V}_i$ is a vertex of the neural mesh $\mathfrak{N}_i$.
The corresponding feature distance of two vertices could be calculated by using the vertex distance $D^{vertex}_{mn}(R)$,
\begin{equation}
    D^{feat}_m(R) = \theta_m - \sum_{V_k \in \mathcal{V}_n}(\frac{\theta_k}{D^{vertex}_{mn}(R)}),
\end{equation}
where $D^{feat}_m(R)$ is the corresponding feature distance of two vertices under the rotation $R$. $\theta_m \in \Theta_j$ is the feature vector of $V_j$ and $\theta_n \in \Theta_i$ is the feature vector of $V_i$. Hence the optimal $R^*$ could be found by minimizing the following equation,
\begin{equation}
    R^* = \underset{R}{\arg\min} \sum_{\theta_m \in \Theta_j} D^{feat}_m(R).
\end{equation}
Then we exploit the optimal $R^*$ to calculate the feature distance $D^{feat}_m(R^*)$ between two meshes and merge two meshes when the distance $D^{feat}_m(R^*) < \tau_{merge}$. $\tau_{merge}$ is the merging threshold and we set $\tau_{merge}$ to 0.8 in this study. The feature vector of each vertex $\theta^*_m \in \Theta^*$ after merging could be calculated by the following equation,
\begin{equation}
    \theta^*_m = \frac{1}{2}\Big(\theta_m + \sum_{V_k \in \mathcal{V}_n}(\frac{\theta_k}{D^{vertex}_{mn}(R^*)})\Big).
\end{equation}
We also exploit the image set $\mathcal{I}_{i}$ that is used to train the mesh $\mathfrak{N}_i$ to train the anchor mesh  $\mathfrak{N}_j$. Specifically, let $P_n$ denote the image pose $I_n \in \mathcal{I}_i$, we train the anchor mesh  $\mathfrak{N}_j$ with the image $I_n$ and the corresponding new pose $P^{'}_n$,
\begin{equation}
    P^{'}_n = P_n R^*.
\end{equation}
During training, we randomly select an anchor mesh and merge the neural meshes by exploiting the above strategy and obtain the category-level feature vectors $\Theta^*$.
% In the following section, we introduce how to exploit the anchor mesh to predict the object pose on the novel image.

\subsection{Evaluation}
\label{sec:eval}
As shown in \figref{fig:overview_eval}, after training, we exploit the feature extractor $\Phi_w(I_m)$ and the anchor neural mesh with the merged feature vectors $\Theta^*$ for estimating the 3D object pose on a novel image. 
We follow previous works \cite{21iclr/wang_nemo,21nips/wang_nvsm} and implement 3D pose estimation in a render-and-compare manner. 
Specifically, given a test image $I_{test}$, we extract its feature maps by using the feature extractor $F_{test} = \Phi(I_{test})$. 
At the same time, we initialize the 3D pose prediction $P$ by sampling a set of preset initial pose. Then we optimize the 3D pose $P$ by exploiting a differentiable renderer $\Re$ to synthesize feature maps $F_{syn}=\Re(\mathfrak{N}, P)$ and computing a feature reconstruction loss between $F_{syn}$ and $F_{test}$. We conduct the gradient optimization of $P$ iteratively. Once the optimization is done, we record the optimized $P$ as the final pose of the test image.
% of the neural mesh with the learned mesh feature vectors $\Theta^*$. The final 3D pose $P$ is determined when the distance between $F_{test}$ and the synthesized $F_{P}$ is minimized.

\begin{table}[!t]
\centering
\resizebox{\linewidth}{!}{%
\begin{tabular}{l|cc|cc}
\toprule
\multirow{2}{*}{Method} & \multicolumn{2}{c|}{1} & \multicolumn{2}{c}{7} \\
                        & Med. $\downarrow$ & Acc$_{10} \uparrow$        & Med. $\downarrow$ & Acc$_{10} \uparrow$  \\ \midrule
NeMo \cite{21iclr/wang_nemo} &   \NA & \NA   & 73.63  & 11.76        \\
NVSM \cite{21nips/wang_nvsm} &   \NA & \NA   & 22.07  & 21.18        \\
Ours                    & \textbf{69.43}    & \textbf{25.54}  & \textbf{16.52} & \textbf{36.42}       \\ \bottomrule
\end{tabular}%
}
\caption{Quantitative results on the car category from the KITTI dataset. While other methods need seven annotated images to train, our model outperforms these methods in the Acc$_{10}$ index without any annotations in the training step. }
\label{tab:kitti}
\end{table}

\begin{table}[]
\setlength{\tabcolsep}{3pt}
\centering
\resizebox{\linewidth}{!}{%
\begin{tabular}{c|ccc|ccc}
\toprule
\multirow{2}{*}{N. Obj} & \multicolumn{3}{c|}{car}                   & \multicolumn{3}{c}{bus} \\
 & Med. $\downarrow$ & Acc$_{10} \uparrow$ & Acc$_{30} \uparrow$ & Med. $\downarrow$  & Acc$_{10} \uparrow$ & Acc$_{30} \uparrow$   \\ \midrule
1    & 92.32 & 7.0   & 28.2                & 17.69    & 32.7      & 69.7    \\
20   & 33.28 & 7.5   & 44.6                & 9.10    & 42.9      & 86.3    \\
50   & 20.23 & 17.6  & 65.0                & 7.62    & 68.0      & 85.7    \\
100  & 13.47 & 32.4  & 70.4                & \textbf{7.43}    & 72.0      & 86.8    \\
150  & \textbf{11.26} & \textbf{43.6}  & \textbf{71.7}                & 7.50    & \textbf{72.2}      & \textbf{87.0}    \\\bottomrule
\end{tabular}%
}
\caption{Ablation study on the number of the training objects in the unannotated images. We present the performance of car and bus categories in the PASCAL3D+ dataset. By leveraging multiple views of more objects in the unannotated images, our model learns stronger generalization ability on the category-level pose estimation tasks and reaches better performance on the accuracy and median error indices.}
\label{tab:abl_num_instance}
\end{table}

\section{Experiment}
In this section, we discuss our experimental results. We first present a brief introduction of our experiment setup. Then we discuss the pose estimation results by training with few pose annotations and without pose annotations. In the end, we conduct a series of ablation studies.
\subsection{Experiment Setup}
\paragraph{Dataset.} We evaluated our model on the PASCAL3D+ dataset \cite{14wacv/xiang_pascal3d} and KITTI dataset \cite{13ijrr/geiger_kitti}. For the Pascal3D+ dataset, we followed previous works \cite{2020cvpr/wang_robust,21nips/wang_nvsm,22cvpr/yin_fishermatch} to evaluate the performance on six categories: aeroplane, bicycle, boat, bus, car and motorbike. The images in these categories have a relatively evenly distributed pose regarding the azimuth angle. For the KITTI dataset, we followed previous works \cite{21nips/wang_nvsm} to evaluate the performance on the car category. We cropped the images with the bounding boxes in the datasets to ensure that the objects are located in the center of the images. We also followed the official KITTI protocol to split the dataset into a training set including 2047 images and a testing set including 681 images.

\paragraph{Implementation details.}We used ResNet50 \cite{16cvpr/he_resnet} pretrained on ImageNet \cite{09cvpr/imagenet} as the image encoder.  we evenly sampled $\omega_{az}$ and $\omega_{el}$ at the fixed 15-degree distance from -90 to 90 degrees. We generated the multiple views of the objects in the training images by using \zeroOneTwoThree~\cite{23iccv/liu_zero123}. As the generated images from \zeroOneTwoThree~have similar scales, we manually adjust a unified camera distance for all the generated images. 
Please refer to \supp~for more implementation details.

\paragraph{Evaluation.} The 3D object pose estimation task predicts three rotation parameters: azimuth, elevation, and in-plane rotation. We follow previous works \cite{21iclr/wang_nemo,21nips/wang_nvsm,22cvpr/yin_fishermatch} and suppose the object scale and center are known. We calculate the pose estimation error between the predicted rotation matrix and the ground truth: $\Delta (P_{pred}, P_{gt}) = \frac{||log m(P_{pred}^TP_{gt})||_F}{\sqrt{2}}$. We report accuracy on two thresholds $\frac{\pi}{6}$ (ACC$_{30}$) and $\frac{\pi}{18}$ (ACC$_{10}$) and the median error (Med.) of the prediction.
\begin{table}[]
\setlength{\tabcolsep}{3pt}
\centering
\resizebox{\linewidth}{!}{%
\begin{tabular}{c|ccc|ccc}
\toprule
\multirow{2}{*}{N. View} & \multicolumn{3}{c|}{car}                   & \multicolumn{3}{c}{motorbike} \\
 & Med. $\downarrow$ & Acc$_{10} \uparrow$ & Acc$_{30} \uparrow$ & Med. $\downarrow$  & Acc$_{10} \uparrow$ & Acc$_{30} \uparrow$   \\ \midrule
1    & 52.13 & 4.3   & 24.4                & 70.53   & 8.2      & 26.8    \\
5    & 25.81 & 6.2   & 59.5                & 47.67   & 11.6      & 37.7    \\
10   & 13.75 & 30.3  & 66.9                & 31.29   & 11.1     & 49.0    \\
20   & \textbf{12.36} &\textbf{ 38.3}  & \textbf{67.9}               & \textbf{23.77}   & \textbf{15.5}     & \textbf{56.9}  \\\bottomrule
\end{tabular}%
}
\caption{Ablation study on the number of training views of each object. We report the pose estimation results on the car and motorbike categories in the PASCAL3D+ dataset. Our model benefits from more object views to learn a better pose correspondence and hence achieve better results on the median error and accuracy indices.}
\label{tab:abl_num_view}
\end{table}

\begin{figure*}[!t]
  \centering
%   \fbox{\rule{0pt}{2in} \rule{0.9\linewidth}{0pt}}
   \includegraphics[width=0.97\linewidth]{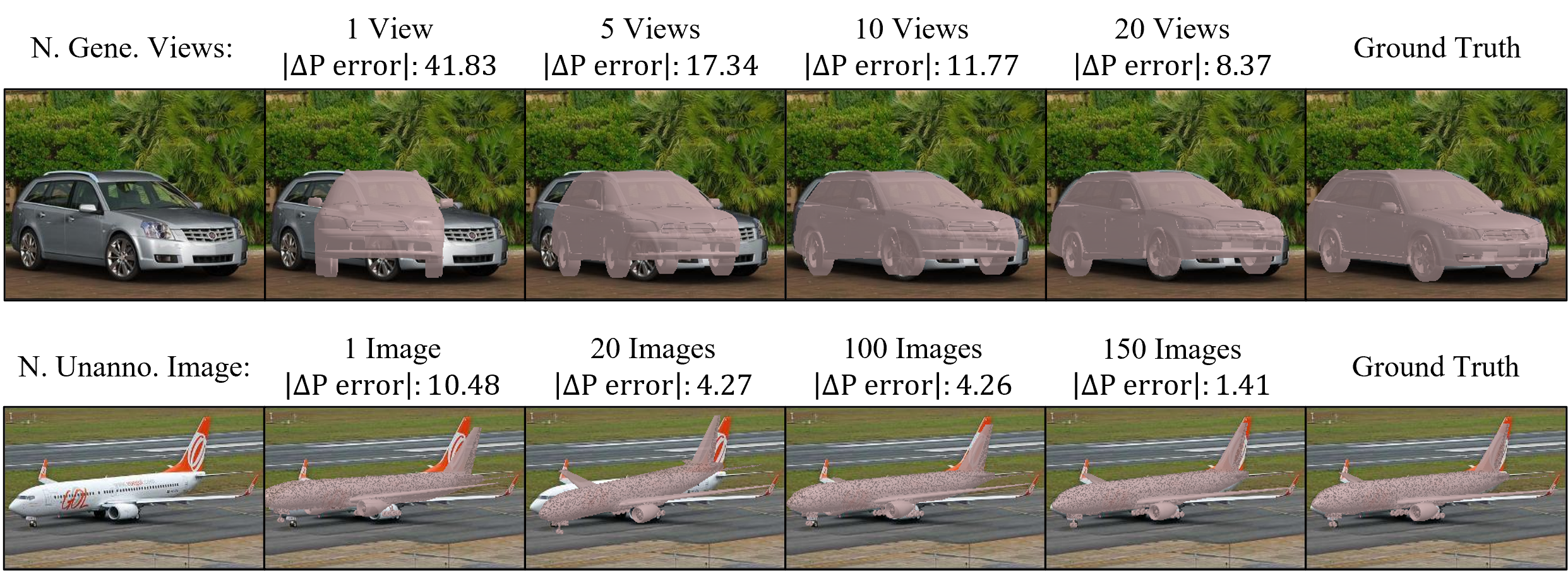}

   \caption{Ablation study on the number of generated views (\textbf{top}) and unannotated images (\textbf{bottom}). We present the results with the corresponding CAD models for better visualization. Our model learns stronger pose correspondence by increasing the object views generated by the diffusion model. Moreover, our model could learn stronger generalization ability by training with more objects from unannotated images.}
   \label{fig:ablation}
\end{figure*}

\subsection{Training with Few Pose Annotations}
As previous works \cite{21nips/wang_nvsm,22cvpr/yin_fishermatch} only trained the category-level pose estimator in the few-shot settings, in this section we tested the performance of our model by training with few annotated images for a fair comparison.
Specifically, we exploited the annotated images to train the anchor neural mesh described in \secref{sec:joint_learn} while jointly optimizing other neural meshes with unannotated images. We reported the pose estimation results on the PASCAL3D+ and KITTI datasets.
\paragraph{Results on the PASCAL3D+ dataset.} For each object category, we follow previous works \cite{21nips/wang_nvsm,22cvpr/yin_fishermatch} to train our model with 7, 20 and 50 object images with annotated poses. We also list the performances of a series of \sArt~methods \cite{21iclr/wang_nemo,21nips/wang_nvsm,22cvpr/yin_fishermatch,16cvpr/he_resnet,18eccv/shou_starmap} in \tabref{tab:pascal3d}. We evaluated StarMap \cite{18eccv/shou_starmap}, NeMo \cite{21iclr/wang_nemo} and FisherMatch \cite{22cvpr/yin_fishermatch} as our baselines as they are recently proposed approaches for few-shot 3D pose estimation. We also implemented a baseline that considers the object pose estimation problem as a classification task as mentioned in StarMap \cite{18eccv/shou_starmap}. Concretely, we implemented a category-specific (Res50-Specific) classifier and a non-specific (Res50-Generic) classifier. The former formulates the object pose estimation problem for all categories as one single classification task and the latter learns a classifier for each category. \tabref{tab:pascal3d} shows that our proposed approach achieves great improvement compared to other \sArt~methods. 
% Specifically, by using seven annotated images, our model improves the ACC$_{30}$ by 3.9\% and median error by 2.0\%.

\paragraph{Results on the KITTI Dataset.} \tabref{tab:kitti} presents the pose estimation results on the KITTI dataset.  We compare our model with NeMo \cite{21iclr/wang_nemo} and NVSM \cite{21nips/wang_nvsm} since they are two of the most competitive methods in this setting. In addition, since the keypoint annotations are not provided in the KITTI dataset, StarMap \cite{18eccv/shou_starmap} cannot be trained on the KITTI dataset without the annotations. \tabref{tab:kitti} shows that our method outperforms other \sArt~methods by a large margin. It demonstrates that our model learns a strong object pose estimator from few annotated images.

\subsection{Training without Pose Annotations}
In this section, we conduct a series of experiments of training an object pose estimator without pose annotations. Specifically, for each category, we trained our model by using multiple views of objects and predicted object poses on the evaluation dataset with the pre-trained model. To evaluate the performance, we exploited one image with the annotated pose to make an alignment of coordinate systems between the predicted poses and the ground truth poses. It is worth noting that previous works \cite{21iclr/wang_nemo,21nips/wang_nvsm,09cvpr/imagenet,22cvpr/yin_fishermatch} fail to predict poses due to the lack of annotated data. We tested the performance of our model on the PASCAL3D+ dataset in \tabref{tab:pascal3d} and the KITTI dataset in \tabref{tab:kitti}.
\paragraph{Performance on the PASCAL3D+ dataset.} We present the visualization results on \figref{fig:qualitative_wo_pose}.
In \tabref{tab:pascal3d}, even without any pose annotations for training, our model achieves 29.3 on median error, which is comparable with the \sArt~performance on the few-shot setting with 7 annotated poses. Moreover, our model could reach 58.8\% performance on the Acc$_{30}$ matrix, which significantly outperforms other \sArt~methods on the seven-annotated setting. Both \figref{fig:qualitative_wo_pose} and \tabref{tab:pascal3d} demonstrate that our model could learn a strong 3D object pose correspondence from multiple views of objects.

\paragraph{Performance on the KITTI dataset.}
In \tabref{tab:kitti}, we present the performance of our model by training without any annotations on the KITTI dataset. Our model achieves 69.43 performance on the median error and 25.54 on Acc$_{10}$. Since the KITTI dataset contains the street scene in the autonomous driving setting, the cropped car images are smaller than the ones on the PASCAL3D+ dataset. In this way, the image quality of the novel object view generated from the \zeroOneTwoThree~\cite{23iccv/liu_zero123} is much lower than the one in the PASCAL3D+ dataset, which leads to a higher median error index. However, even without the object poses, our method reaches significantly better performance than the \sArt~methods on the seven-shot setting. It demonstrates the effectiveness of our model by learning a strong 3D pose correspondence from multiple object views.

\subsection{Ablation studies}
In this section, we conduct a series of ablation studies of our model by training our model without pose annotations.
We first test the performance of our model with different numbers of unannotated images in the training pipeline. Then we conduct the ablation study on the number of object views. 

\paragraph{Number of the unannotated images.} Since our model exploits multiple neural meshes for training with multiple views of the objects in the unannotated images, we test the model performance by using different numbers of unannotated images. We present the results in \figref{fig:ablation} and \tabref{tab:abl_num_instance}. When training our model only with one neural mesh and the multiple views from one object image, \figref{fig:ablation} shows that the predicted pose has a relatively large pose error and the performance of Acc$_{10}$ and Acc$_{30}$ on the car and bus categories in the PASCAL3D+ dataset has a significant drop, which demonstrates that the category-level generalization ability can hardly be learned from only one object in the category. With the increase of unannotated images, the performance of Acc$_{10}$ and Acc$_{30}$ indices gradually increase, which demonstrates that the category-level generalization ability of our model could be obtained by jointly learning from multiple views of hundreds of objects in the unannotated images.

\paragraph{Number of the object views.}
As different views of an object provide different details of 3D pose correspondence, we evaluate the correspondence learning ability  by training with different numbers of object views. In \tabref{tab:abl_num_view} and \figref{fig:ablation}, we present the results on the PASCAL3D+ dataset. Since the 3D pose correspondence cannot learn from one view of an object, there is a performance drop when training the model with only one view of each object. By increasing the training views of objects, our model gradually reaches higher performances on  Acc$_{10}$ and Acc$_{30}$ indices and has lower median errors of pose estimation. It shows that the multiple object views significantly contribute to the pose estimator learning.

\section{Conclusion}

In this paper, we propose to learn an object pose estimator without pose annotations. By leveraging the diffusion model \ie~\zeroOneTwoThree~to generate multiple views of objects, we propose to learn the pose estimator from the generated view images. As the generated images from the diffusion model have both image quality issues and severe noise on the pose labels, we propose to exploit an image encoder to extract the image features of different image views and learn the pose correspondence from the extracted image features. Experiments on the PASCAL3D+ and KITTI datasets demonstrate the effectiveness of our model.

\noindent \textbf{Acknowledgments.} We gratefully acknowledge support via Army Research Laboratory award W911NF2320008 and ONR with award N00014-21-1-2812.
% \appendix

% \section*{Ethical Statement}

% There are no ethical issues.

% \section*{Acknowledgments}

% The preparation of these instructions and the \LaTeX{} and Bib\TeX{}
% files that implement them was supported by Schlumberger Palo Alto
% Research, AT\&T Bell Laboratories, and Morgan Kaufmann Publishers.
% Preparation of the Microsoft Word file was supported by IJCAI.  An
% early version of this document was created by Shirley Jowell and Peter
% F. Patel-Schneider.  It was subsequently modified by Jennifer
% Ballentine, Thomas Dean, Bernhard Nebel, Daniel Pagenstecher,
% Kurt Steinkraus, Toby Walsh, Carles Sierra, Marc Pujol-Gonzalez,
% Francisco Cruz-Mencia and Edith Elkind.

\bibliographystyle{named}
\bibliography{ijcai24}

\end{document}

\maketitle

\begin{table*}[!hbt]
\centering
\tabFormat
\resizebox{\textwidth}{!}{%
\begin{tabular}{l|c|c|c|c|c|c}
\toprule
GT number  & bus            & car             & boat     & motorbike      & bicycle       & aeroplane     \\ \midrule
1 instance                  & 87.4 / 75.0 / 6.80 & 70.4 / 32.4 / 13.47 & 33.6 / 6.5 / 64.60  & 60.3 / 13.5 / 22.37 & 37.1 / 2.9  / 64.78 & 55.5 / 9.4 / 26.12  \\
7 instances                 & 87.4 / 65.4 / 7.41 & 78.1 / 53.1 / 9.49  & 38.0 / 9.5 / 48.35  & 54.0 / 7.7 / 27.44  & 47.0 / 5.9 / 32.18  & 35.3 / 4.4 / 55.48  \\
20 instances                & 88.7 / 60.0 / 8.39 & 87.6 / 58.9 / 8.76  & 35.1 / 10.0 / 64.69   & 57.7 / 11.1 / 25.39 & 54.7 / 12.1 / 26.30 & 60.0 / 21.9 / 20.34 \\
50 instances                & 90.2 / 80.1 / 4.80 & 93.7 / 69.8 / 7.16  & 46.3 / 14.2 / 35.28 & 74.6 / 16.2 / 19.75 & 68.1 / 16.9 / 20.61 & 67.0 / 27.1 / 16.56 \\\bottomrule
\end{tabular}%
}
\caption{Detailed results on the PASCAL3D+ Dataset. We report the results on seven classes: bus, car, boat, motorbike, bicycle and aeroplane. We list the  Acc$_{30} \uparrow$ / Acc$_{10} \uparrow$ / Med. $\downarrow$ scores in the table. We show the pose estimation results by training without annotations in the first result row, which only uses one instance for each object category as a pose definition in the evaluation set. The rows of 7, 20 and 50 instances present the results in the corresponding few-shot settings. }
\label{tab:details_pascal3d}

\end{table*}

\section{Code}

We will publicly share our implementation codes in the future.

\section{Implementation Details about Diffusion Model}
We used \zeroOneTwoThree~\cite{23iccv/liu_zero123} to generate multiple object views. The related poses of objects that are introduced into \zeroOneTwoThree~as input are listed in \tabref{tab:diffusion_pose}. We totally generate $12\times7=84$ views for each object. During training, we use 84 views with the origin images of each object to optimize the neural meshes.
\begin{table}[]
\centering
\resizebox{0.8\linewidth}{!}{%
\begin{tabular}{c|c}
\toprule
$\omega_{az}$ in degree   & 0, $\pm$15, $\pm$30, $\pm$45, $\pm$60, $\pm$90, 180 \\
\midrule
$\omega_{el}$ in degree & 0, $\pm$15, $\pm$30, $\pm$45    \\ \bottomrule
\end{tabular}%
}
\caption{The object pose inputs for the diffusion model. we report the used 12 different $\omega_{az}$ and 7 different $\omega_{el}$ for generating different views of object. The total number of generated views for each object is $12\times7=84$.}
\label{tab:diffusion_pose}
\end{table}

\section{Detailed Results on the PASCAL3D+ dataset}
We show the detailed results on the PASCAL3D+ dataset \cite{14wacv/xiang_pascal3d} in \tabref{tab:details_pascal3d}.

% \appendix

% \section*{Ethical Statement}

% There are no ethical issues.

% \section*{Acknowledgments}

% The preparation of these instructions and the \LaTeX{} and Bib\TeX{}
% files that implement them was supported by Schlumberger Palo Alto
% Research, AT\&T Bell Laboratories, and Morgan Kaufmann Publishers.
% Preparation of the Microsoft Word file was supported by IJCAI.  An
% early version of this document was created by Shirley Jowell and Peter
% F. Patel-Schneider.  It was subsequently modified by Jennifer
% Ballentine, Thomas Dean, Bernhard Nebel, Daniel Pagenstecher,
% Kurt Steinkraus, Toby Walsh, Carles Sierra, Marc Pujol-Gonzalez,
% Francisco Cruz-Mencia and Edith Elkind.

\bibliographystyle{named}
\bibliography{ijcai24}